\documentclass[10pt,twocolumn,letterpaper]{article}

\usepackage{cvpr}      
\definecolor{cvprblue}{rgb}{0.21,0.49,0.74}
\usepackage[pagebackref,breaklinks,colorlinks,allcolors=cvprblue]{hyperref}

\usepackage[table]{xcolor}
\usepackage{float}
\usepackage{graphicx}    
\usepackage{booktabs}    
\usepackage{multirow}    
\usepackage{amsmath}     
\usepackage{subcaption}  
\usepackage{caption}
\usepackage{pifont}
\usepackage{bm}
\definecolor{best}{HTML}{FFE6E6}   
\definecolor{second}{HTML}{FFF2E6} 
\definecolor{third}{HTML}{FFFFCC}  

\newcommand{\method}{SLARM}

\def\paperID{6478} 
\def\confName{CVPR}
\def\confYear{2026}

\title{%
  SLARM: \textbf{S}treaming and \textbf{L}anguage-\textbf{A}ligned \textbf{R}econstruction \textbf{M}odel \\
  for \textbf{D}ynamic \textbf{S}cenes%
}

\author{
  Zhicheng Qiu$^{\star \dagger}$,
  Jiarui Meng$^{\star}$,
  Tong-an Luo$^{\star}$,
  Yican Huang,
  Xuan Feng,
  Xuanfu Li$^{\ddagger}$,
  Zhan Xu \\
  {\textit{Huawei Technologies Ltd.}} \\
  \footnotesize \texttt{\{chiu.chih.cheng, mengjiarui, luotongan, huangyican, fengxuan18, lixuanfu, zhan.xu\}@huawei.com} \\
  \small{$^{\star}$ denotes equal contribution, $^{\dagger}$ denotes project lead, $^{\ddagger}$ denotes corresponding author}
}

\begin{document}

\twocolumn[{
      \renewcommand\twocolumn[1][]{#1}
      \maketitle
      \centering
      \vspace{-0.6cm}
      \includegraphics[width=\textwidth]{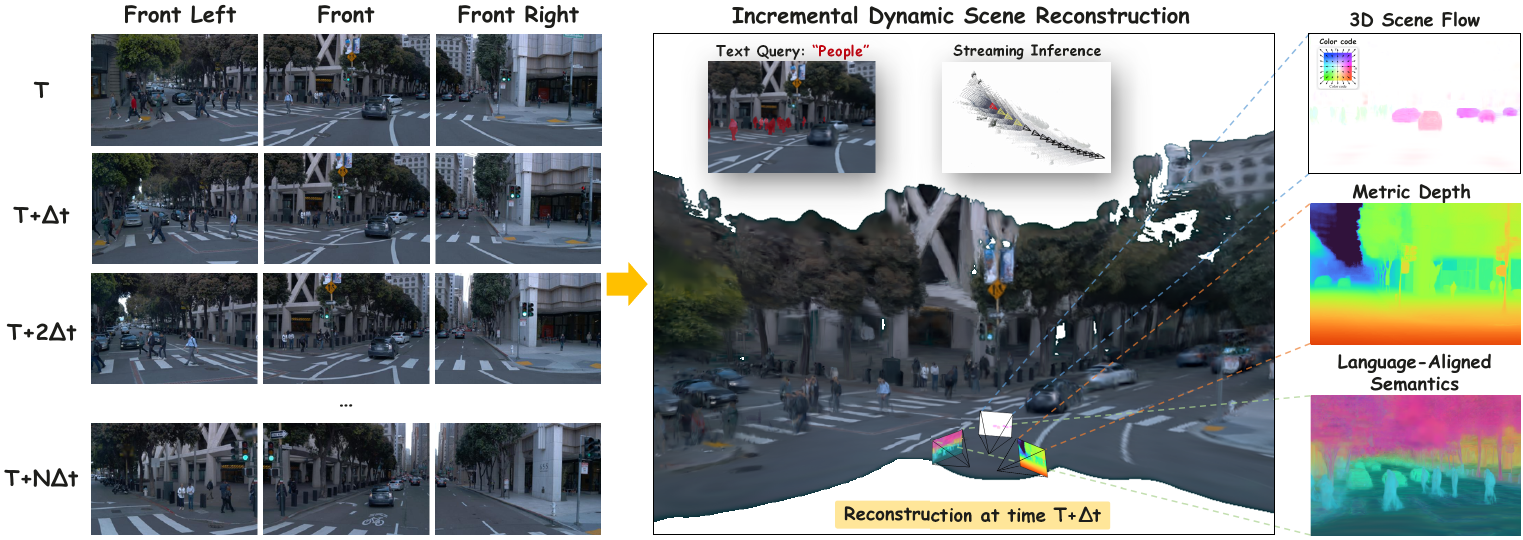}
      \vspace{-0.6cm}
      \captionsetup{type=figure}
      \caption{\textbf{SLARM} is a large feedforward Transformer using a self-supervised approach for fast and accurate inference of 3D scene flow, metric depth and language-aligned semantics in dynamic scenes. For real-time inference deployment in autonomous driving and embodied AI applications, our model also supports incremental streaming inference.}
    }
  \label{fig:demo}
  \vspace{0.5cm}]

\begin{abstract}
    We propose \textbf{\method}, a feed-forward model that unifies \textbf{dynamic scene reconstruction}, \textbf{semantic understanding}, and \textbf{real-time streaming inference}. SLARM captures complex, non-uniform motion through higher-order motion modeling, trained solely on differentiable renderings without any flow supervision. Besides, SLARM distills semantic features from LSeg to obtain language-aligned representations. This design enables semantic querying via natural language, and the tight coupling between semantics and geometry further enhances the accuracy and robustness of dynamic reconstruction. Moreover, SLARM processes image sequences using window-based causal attention, achieving stable, low-latency streaming inference without accumulating memory cost. Within this unified framework, SLARM achieves state-of-the-art results in dynamic estimation, rendering quality, and scene parsing, improving motion accuracy by \textbf{21\%}, reconstruction PSNR by \textbf{1.6 dB}, and segmentation mIoU by \textbf{20\%} over existing methods.
\end{abstract}
\section{Introduction}
The advent of neural radiance fields (NeRF)~\cite{nerf} and 3D Gaussian Splatting (3DGS)~\cite{3dgs} has significantly advanced the efficiency and practicality of 3D scene reconstruction. Building on these advances, research has extended to dynamic scene reconstruction~\cite{nerf-ds, dnerf,nerf-ds, dynamic3dgs, 4dgs, deformabl3dgs, yan2025instant, gao2024hicom, fu2025recon}, enabling new possibilities for immersive interaction and experience. However, existing approaches typically require optimization times ranging from several minutes to several hours and are often limited to per-scene overfitting, with limited ability to generalize across scenes.

Recently, the rise of feed-forward models such as DUST3R~\cite{dust3r} and PixelSplat~\cite{pixelsplat} has brought data-driven paradigms into 3D reconstruction. In particular, works like VGGT~\cite{vggt} and MapAnything~\cite{mapanything} have challenged the conventional optimization-based reconstruction pipeline, further evolving into general-purpose 3D foundation models. Nevertheless, current approaches remain largely confined to static scenes, and feed-forward reconstruction of dynamic environments remains underexplored.

A recent work STORM~\cite{storm} enables dynamic 3D scene reconstruction from multi-timestep, posed images. However, it suffers from several key limitations: (1) \textbf{oversimplified motion modeling}—it assumes constant-velocity motion, which fails to capture the complex, nonlinear dynamics prevalent in real-world scenes; (2) \textbf{limited functionality}—it focuses solely on geometric reconstruction and lacks high-level semantic understanding, hindering its applicability to downstream perception and reasoning tasks; (3) \textbf{inefficient inference}—it requires batching multiple input frames and relies on interpolation across timesteps, lacking the capability for incremental, streaming inference.

To address these shortcomings, we propose \textbf{SLARM}, a unified 4D Gaussian reasoning framework that simultaneously achieves \textbf{dynamic reconstruction}, \textbf{semantic understanding}, and \textbf{streaming inference}. Our main contributions are summarized as follows.

\noindent \ding{113} \textbf{Accurate and Efficient Motion Modeling}:
We propose a motion representation based on a higher-order motion function, enabling effective modeling of non-uniform motion without explicit supervision and significantly improving geometric and dynamic fidelity.

\noindent \ding{113} \textbf{Language-Aligned 4D Semantics}: We distill language-aligned semantic knowledge from 2D foundation model LSeg into the SLARM model, endowing it with text-aligned semantic features that are directly queryable by large language models (LLMs), thereby achieving better dynamic scene understanding and reasoning.

\noindent \ding{113} \textbf{Streaming Inference Architecture}: We process each frame independently and incorporate window-based attention to achieve constant low-latency streaming inference—eliminating the need for batched or sliding-window processing and supporting long-horizon, low-latency deployment in real-world scenarios such as autonomous driving and embodied AI.

\noindent \ding{113} \textbf{Unified Multi-Task Learning}: We jointly optimize geometry, motion, and semantics within a single forward pass, enabling mutual task enhancement and outperforming specialized methods in reconstruction fidelity, motion accuracy, and semantic alignment.

\section{Related Work}

\textbf{4D Dynamic Scene Reconstruction.}
Early dynamic modeling efforts incorporated temporal information into NeRF \cite{dnerf, nerfies, TiNeuVox, hnerf} but suffered from slow training and rendering. Recent works leverage 3D Gaussians for efficiency: 4DGT~\cite{4dgt} learns a 4D Gaussian transformer from monocular videos but requires dense 3D supervision; MoVieS~\cite{movies} achieves fast dynamic view synthesis but depends on precomputed 4D point trajectories or multi-view stereo; STORM~\cite{storm} models outdoor dynamics via Gaussian motion but assumes linear motion, failing on complex non-rigid motions. Other dynamic approaches \cite{easi3r, monst3r, d2ust3r, st4rtrack, megasam, geometrycrafter, geo4d, stereo4d} either rely on strong supervision, impose restrictive motion priors, or are constrained by specific inputs. SLARM requires only 2D renderings and camera poses, implicitly learning scene flow through temporal consistency of Gaussian attributes (position, covariance, opacity). It uses learned, non-parametric temporal deformations to capture arbitrary motion patterns without restrictive priors.
Pi3~\cite{pi3} addresses dynamics via permutation-equivariant geometry learning but only produces per-frame reconstructions, lacking inter-frame correspondences for motion reasoning. In contrast, SLARM explicitly models temporal Gaussian motion to yield dense, differentiable flow fields for downstream control.
Novel view synthesis methods \cite{anysplat, lvsm, matrix3d, flare, fillerbuster} have focused on photorealism or generation to the exclusion of dynamic 3D modeling.

\noindent
\textbf{Semantic 3D Understanding and Language Alignment.} Semantic 3D reconstruction has advanced with vision-language models. LERF~\cite{lerf} pioneered distilling 2D CLIP~\cite{clip} features into 3D radiance fields. With 3DGS, \cite{feature3dgs, langsplat, semanticsplat, opengaussian, legaussian, gaussiangrouping, legs, hybridf} improved semantic-geometric modeling but relied on per-scene optimization or added computational overhead.
Efficient feed-forward methods \cite{pe3r, lsm, gsemsplat} had limitations in reconstruction quality or robustness. Uni3R~\cite{uni3r} unified static 3D reconstruction and language-aligned semantics via 3DGS but lacked temporal modeling.
SLARM extends semantic capability to 4D scenes by distilling LSeg features into time-consistent Gaussian descriptors, enabling language-queryable dynamics. Unlike closed-set~\cite{semanticnerf, mask3d} methods, it enables zero-shot generalization and seamless LLM integration for high-level reasoning in dynamic scenes.

\noindent
\textbf{Streaming and Real-Time 3D Perception.} Real-time 3D reconstruction is critical for robotics and autonomous systems.
SLAM-based methods \cite{monoslam, slam3r, vggt-slam} prioritize accuracy but face latency and speed-precision trade-offs; memory pool-based approaches \cite{spann3r, driv3r, point3r} struggle with temporal fusion or dynamic scene robustness; sequential models \cite{cut3r, streamvggt, stream3r} suffer from long-range dependency limitations or dynamic adaptability issues.
SLARM overcomes these via a pure streaming paradigm: it processes each frame independently while propagating a compact hidden state, ensuring constant latency and memory usage.

\begin{figure*}[t]
    \centering
    \includegraphics[width=0.95\linewidth]{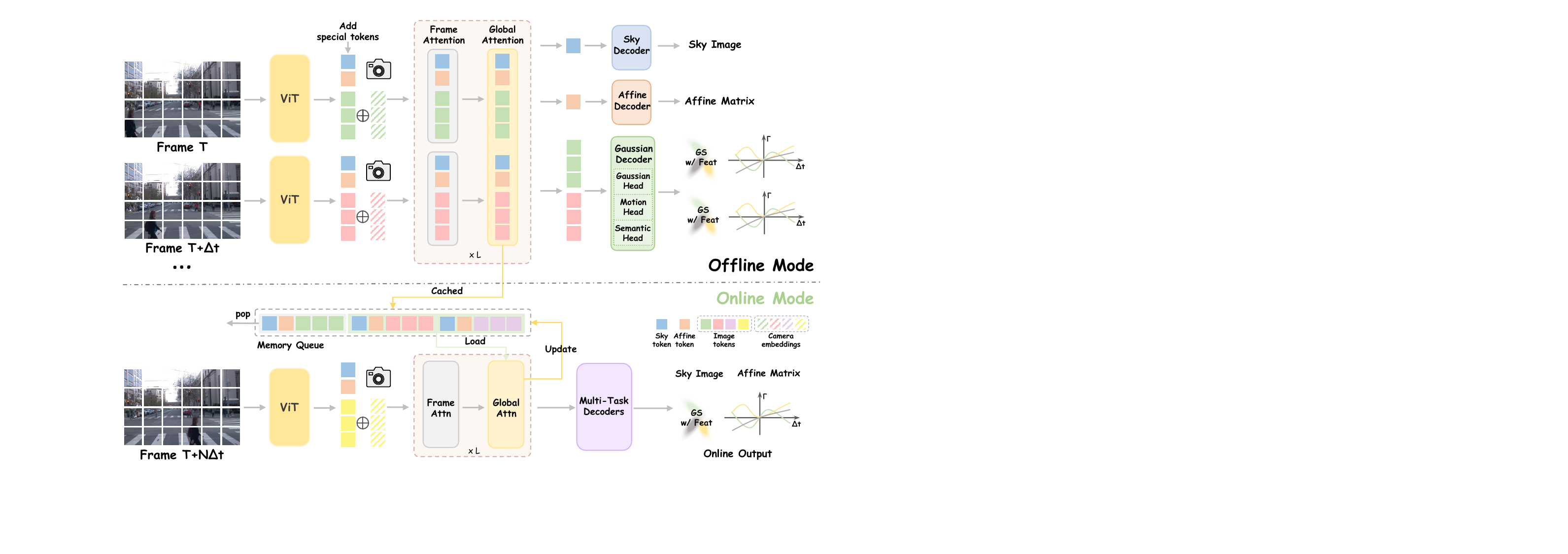}
    \caption{Pipeline of our SLARM model for streaming 4D Gaussian reconstruction. The SLARM model begins by extracting image tokens using a shared-weight Vision Transformer (ViT), and then appending special tokens to these image tokens. These tokens then undergo Frame Attention and Global Attention mechanisms, before being fed into the corresponding decoders to output the parameters.}
    \label{fig:pipeline}
    \vspace{-5mm}
\end{figure*}

\section{Method}

\textbf{Task Definition.}
Given a video sequence $\{\mathbf{I}_t \in \mathbb{R}^{H \times W \times 3} \}_{t=1}^T$ with known camera intrinsics $\mathbf{K}_t \in \mathbb{R}^{3 \times 3}$ and extrinsics $\mathbf{P}_t \in \mathbb{SE}(3)$, our goal is to achieve incremental dynamic scene reconstruction using a feedforward model. To this end, we propose \textbf{SLARM}, which maintains an explicit \textbf{4D Gaussian Splatting (4DGS)} representation at each time step $t$ that simultaneously: (i) reconstructs current geometry and appearance; (ii) encodes per-Gaussian 3D scene flow; and (iii) associates each Gaussian with a language-aligned semantic feature for text-based querying.
This formulation addresses three core challenges: unsupervised scene flow learning, vision--language grounding of geometric primitives, and constant-latency, history-aware incremental updates.

\subsection{SLARM}

Our method, \textbf{SLARM}, follows the pipeline illustrated in Figure~\ref{fig:pipeline}. Given a video sequence $\{\mathbf{I}_t\}_{t=1}^{T}$, we first extract visual features using a Vision Transformer (ViT), where each frame $\mathbf{I}_t$ is partitioned into non-overlapping patches. To inject geometric priors, we encode each pixel's viewing ray as a 6D Pl\"ucker coordinate~\cite{plucker1865} derived from known camera intrinsics and extrinsics, linearly project it, and add it to the corresponding visual tokens. Temporal context is incorporated through a learnable embedding of the absolute timestamp $t$~\cite{scalable_diffusion_transformers}. Following STORM~\cite{storm}, we further concatenate two specialized tokens: a \textit{Sky token} to model the background sky region and an \textit{Affine token} to compensate for exposure and white-balance variations across multi-view cameras. The resulting enriched token sequence is then processed by an \textit{Alternating-Attention Transformer} backbone~\cite{vggt}, which alternates between frame-wise and global self-attention to effectively capture spatio-temporal structure.

Gaussian Decoder regresses a \textbf{pixel-aligned 4DGS} representation: for each pixel, it predicts 3D position $\bm{\mu} \in \mathbb{R}^3$, rotation $\bm{q} \in \mathbb{R}^4$, scale $\bm{s} \in \mathbb{R}^3$, opacity $\alpha \in [0,1]$, and color $\bm{c} \in \mathbb{R}^3$. The position is computed as $\bm{\mu} = \bm{o} + d \cdot \bm{r}$, where $d \in \mathbb{R}$ is predicted depth and $\bm{o}, \bm{r} \in \mathbb{R}^3$ are the ray origin and direction derived from the camera pose. Two auxiliary heads further predict: (i) a dynamic attribute vector encoding scene flow; and (ii) a semantic feature vector aligned with a vision–language embedding space. All outputs are per-frame and pixel-aligned, enabling direct 2D–3D correspondence and incremental reconstruction. More model architecture details are in the \textcolor{red}{\textbf{Supplementary Material}}.

We next detail our approach along three aspects: (1) self-supervised dynamic modeling without ground truth scene flows; (2) language-aligned semantics via distillation from 2D foundation models; and (3) streaming inference via incremental state propagation.

\subsection{Dynamic Representation and Learning}
\label{sec:3.2}

Previous optimization-based approaches to dynamic scene modeling commonly formulate motion using time-dependent displacement field, which requires precise estimation of the object's position at each individual timestamp, resulting in low prediction accuracy in real scene. To alleviate this requirement, STORM~\cite{storm} adopts a velocity-based formulation by predicting instantaneous object velocities. However, it relies on the assumption of constant velocity over time, which is insufficient to model complex, non-uniform motion patterns such as those observed in human limbs, as shown in Figure~\hyperref[fig:qualitative_1]{\ref*{fig:qualitative_1}(a)}.

\vspace{0.5em}
\noindent\textbf{High-Order Motion Modeling.}
To overcome these limitations, we model displacement as a differentiable function of time using a multi-order Taylor expansion. Specifically, for each order $l \in \{0, \dots, L-1\}$, the network predicts a scalar speed $s_l \in \mathbb{R}$ and a 3D directional vector $\bm{v}_l \in \mathbb{R}^3$. The directional vector is L2-normalized, and the resulting motion coefficient $\bm{m}_l \in \mathbb{R}^3$ is computed as
\begin{equation}
    \bm{m}_l = s_l \cdot \frac{\bm{v}_l}{\|\bm{v}_l\|_2}.
\end{equation}

Given a temporal offset $\Delta t$, the total displacement is then obtained by aggregating contributions from all orders in a Taylor-like expansion up to the $L$-th order:
\begin{equation}
    \bm{\Gamma}(\Delta t) = \sum_{l=0}^{L-1} \frac{\bm{m}_l \cdot (\Delta t)^{l+1}}{(l+1)!}.
\end{equation}

We adopt $L = 3$ in our experiments, corresponding to a \textbf{3rd-order} expansion that explicitly models the first three temporal derivatives of position—velocity, acceleration, and jerk—offering a compact yet expressive representation for complex real-world dynamics.

\noindent\textbf{Rendering-Supervised Motion Learning.}
Our approach learns scene motion in a fully self-supervised manner without requiring ground-truth scene flow. Specifically, given an input frame at time $t$ and a supervision frame at time $t + \Delta t$, the network predicts the forward scene flow $\bm{\Gamma}(\Delta t)$ that maps each 3D Gaussian position from time $t$ to $t + \Delta t$.

Let $\mathcal{G}_t = \{ \bm{\mu}_i, \alpha_i, \bm{q}_i, \bm{s}_i, \bm{c}_i \}_{i=1}^N$ denote the set of 3D Gaussians at time $t$. We only allow the positions to evolve over time, keeping all other attributes fixed. The transformed Gaussians at time $t + \Delta t$ are thus:
\begin{equation}
    \mathcal{G}_{t \to t+\Delta t} = \{ \bm{\mu}_i + \bm{\Gamma}_i(\Delta t),\alpha_i, \bm{q}_i, \bm{s}_i, \bm{c}_i \}_{i=1}^N.
\end{equation}

We render this warped Gaussian set to obtain a synthesized image $\hat{\mathbf{I}}_{t + \Delta t}$. The motion prediction is supervised by a combination of pixel-wise MSE and perceptual LPIPS loss against the ground-truth supervision frame $\mathbf{I}_{t + \Delta t}$:
\begin{equation}
    \begin{aligned}
        \mathcal{L}_{\mathrm{rgb}}
         & = \bigl\| \hat{\mathbf{I}}_{t + \Delta t} - \mathbf{I}_{t + \Delta t} \bigr\|_2^2                                                      \\
         & \quad + \lambda_{\mathrm{lpips}} \cdot \operatorname{LPIPS}\bigl( \hat{\mathbf{I}}_{t + \Delta t},\, \mathbf{I}_{t + \Delta t} \bigr),
    \end{aligned}
\end{equation}
where $\lambda_{\mathrm{lpips}}$ is set to 0.05.

\begin{figure*}[t]
    \centering
    \includegraphics[width=0.95\linewidth]{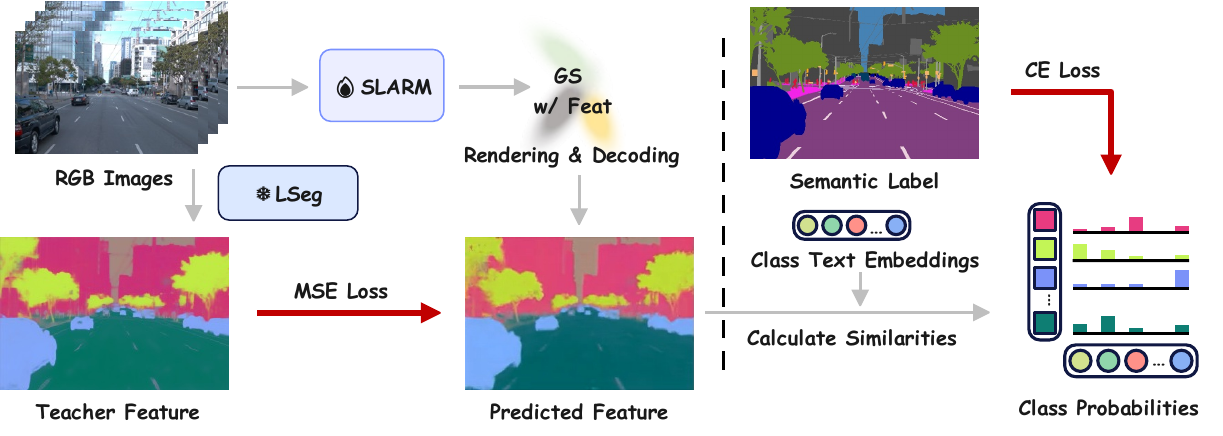}
    \vspace{-3mm}
    \caption{Illustration of semantic supervision in SLARM. Left: Self-supervised feature learning via distillation from LSeg, where rendered semantic Gaussians are decoded. Right: Supervised training on labeled data by aligning predictions with class text embeddings.}
    \label{fig:semantic_pipeline}
    \vspace{-2mm}
\end{figure*}

\begin{figure*}[t!]
    \centering
    \begin{minipage}[b]{0.48\linewidth}
        \centering
        \includegraphics[width=\linewidth]{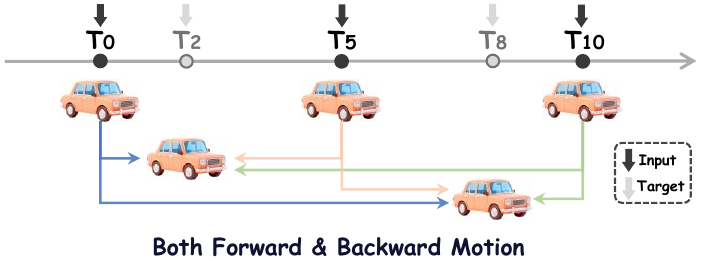}
        \subcaption{Offline Inference}
        \label{fig:motion1}
    \end{minipage}
    \hfill
    \begin{minipage}[b]{0.48\linewidth}
        \centering
        \includegraphics[width=\linewidth]{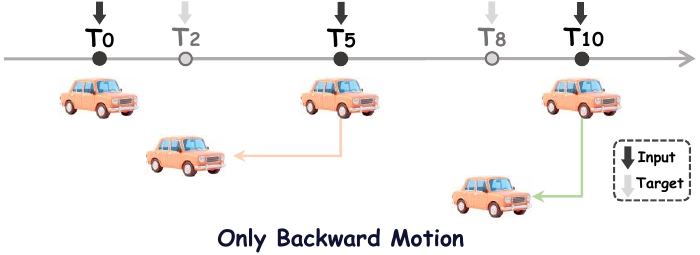}
        \subcaption{Online Inference}
        \label{fig:motion2}
    \end{minipage}
    \vspace{-2mm}
    \caption{Illustration of motion handling for dynamic Gaussians under two modes.
        In offline inference, the target frame is synthesized by interpolating all input frames.
        In online inference, the target frame is reconstructed via backward warping from the subsequent frame.}
    \label{fig:motion_patterns}
    \vspace{-5mm}
\end{figure*}

\subsection{Semantic Feature Distillation}
\label{sec:semantic_distill}

Our approach extends the semantic distillation framework to a dynamic setting, where 2D semantic information is distilled into a \textbf{4DGS} representation. This extension not only enhances the model's ability to understand dynamic scenes but also leverages semantic consistency to improve motion estimation accuracy.

\vspace{0.5em}
\noindent\textbf{Extending 2D semantics to 4D Gaussian primitives.}
Similar to Uni3R~\cite{uni3r}, each Gaussian primitive in our model is augmented with a high-dimensional semantic feature vector $\bm{f}^{\text{sem}}_j \in \mathbb{R}^d$. However, different from the static 3D representation in Uni3R, our Gaussians evolve over time according to the higher-order motion function $\bm{\Gamma}$. During rendering, we synthesize both RGB images and semantic feature maps $\hat{\mathbf{F}}_{t + \Delta t} \in \mathbb{R}^{h \times w \times d}$ at supervision frame at timestamps $t + \Delta t$ by alpha-blending the semantic attributes of time-warped Gaussians.

To supervise this 4D semantic field, we extract per-frame 2D semantic features $\tilde{\mathbf{F}}_{t + \Delta t} \in \mathbb{R}^{h \times w \times c}$ using the frozen 2D foundation model LSeg~\cite{li2022lseg}. The distillation loss is defined as the MSE between rendered and 2D features:
\begin{equation}
    \mathcal{L}_{\text{sem}} = \| \tilde{\mathbf{F}}_{t + \Delta t} - \hat{\mathbf{F}}_{t + \Delta t}' \|_2^2,
\end{equation}
where $\hat{\mathbf{F}}_{t + \Delta t}' \in \mathbb{R}^{h \times w \times c}$ is the decoded semantic map from a lightweight MLP for reducing memory useage.

\vspace{0.5em}
\noindent\textbf{Enhancing semantic reconstruction with semantically annotated data.}

For data with existing semantic annotations, we aim to leverage such data effectively to further enhance semantic reconstruction performance. Since we distill features from LSeg, we opt to utilize semantically annotated data in the same manner as it.

As is illustrated in the right part of Figure~\ref{fig:semantic_pipeline}, reconstructed semantics are decoded into a feature map with width $w$ and height $h$. We then calculate the similarities between each feature $\bm{f}_{ij} \in \mathbb{R}^{c}$ in this feature map and the text feature $\bm{t}_{k} \in \mathbb{R}^{c}$ of each category (extracted using the CLIP text encoder) via the inner product, and convert these similarities into category probabilities via softmax. At this point, this reduces to a classification task, for which we employ cross-entropy loss:

\begin{equation}
    \mathcal{L}_{\text{cls}} = \frac{1}{hw} \sum_{i,j=1}^{h,w} -\log\left( \frac{\exp\left( \bm{f}_{ij} \cdot \bm{t}_{k_{ij}} /\tau \right)}{\sum_{k=1}^{K} \exp\left( \bm{f}_{ij} \cdot \bm{t}_{k}/\tau \right)} \right),
\end{equation}
where $k_{ij}$ denotes the ground-truth category of the feature at position $(i, j)$, and $\tau$ is a user-defined temperature parameter that we set to $0.07$ (the same as that in LSeg).

\subsection{Streaming 4D Scene Reconstruction}
\label{subsec:streaming_4d}

In contrast to streaming methods like StreamVGGT~\cite{streamvggt} and Stream3R~\cite{stream3r}, which reconstruct only per-frame 3D geometry, we tackle \textbf{streaming 4D scene reconstruction}—jointly modeling instantaneous geometry and its continuous temporal deformation under real-time latency and memory constraints. Unlike offline dynamic reconstruction methods that interpolate using both past and future frames, our approach adheres to strict causality: at inference, only current and past observations are available. This requires retroactively refining dynamic content by propagating Gaussian primitives backward in time, while preserving static elements for geometric consistency.

Formally, under the streaming setting, our model outputs both the 3D Gaussian representation $\mathcal{G}_t$ and the associated displacement field $\Delta \bm{\mu}_t$ at time $t$, conditioned exclusively on the observed frames up to the current timestamp:
\begin{equation}
    \big( \mathcal{G}_t, \bm{\Gamma}_t \big) = \phi\left( \mathbf{I}_t \mid \mathbf{I}_{t-\Delta t}, \mathbf{I}_{t-2\Delta t}, \dots \right),
    \label{eq:streaming_output_causal}
\end{equation}
where $\phi$ denotes our causal streaming reconstructor, and $\Delta t$ is the temporal stride (typically $\Delta t = 5$).

As illustrated in Figure~\hyperref[fig:motion_patterns]{\ref*{fig:motion_patterns}(b)}
, the dynamic Gaussians in $\mathcal{G}_t$ are propagated backward only to the most recent historical frame at time $t - \Delta t$. Consequently, the scene representation over the interval $[t - \Delta t, t]$ is composed of two disjoint components:
\begin{equation}
    \mathcal{S}_{[t - \Delta t, t]} =
    \underbrace{\mathcal{G}^{\mathrm{static}}_{t - \Delta t} \ \cup  \ \mathcal{G}^{\mathrm{static}}_{t}}_{\text{static geometry}}
    \ \cup
    \underbrace{\mathcal{G}^{\mathrm{dynamic}}_{t \to t-\Delta t'}}_{\text{backward dynamics}},
    \label{eq:scene_composition}
\end{equation}
where $\mathcal{G}^{\mathrm{static}}_{t - \Delta t'}$ denotes the static subset of Gaussians from timestamp $t - \Delta t'$. Specifically, each Gaussian primitive $g=(\bm{\mu}, \alpha, \bm{q}, \bm{s}, \bm{c}) \in \mathcal{G}$ is associated with a displacement field $\bm{\Gamma}_g $. We partition $\mathcal{G}$ into static and dynamic subsets based on motion magnitude:
\begin{equation}
    \begin{split}
        \mathcal{G}^{\mathrm{static}}  = \left\{ g \in \mathcal{G} \,\middle|\, \|\bm{\Gamma}_g(\Delta t) \leq \tau_m \right\}, \\
        \mathcal{G}^{\mathrm{dynamic}}  = \left\{ g \in \mathcal{G} \,\middle|\, \|\bm{\Gamma}_g(\Delta t) >  \tau_m \right\},
    \end{split}
    \label{eq:static_dynamic_split}
\end{equation}
where $\tau_m > 0$ is a motion threshold. The above design enables streaming inference and prevents rendering holes at new timesteps.

\begin{figure*}[t]
    \centering
    \includegraphics[width=1.0\linewidth]{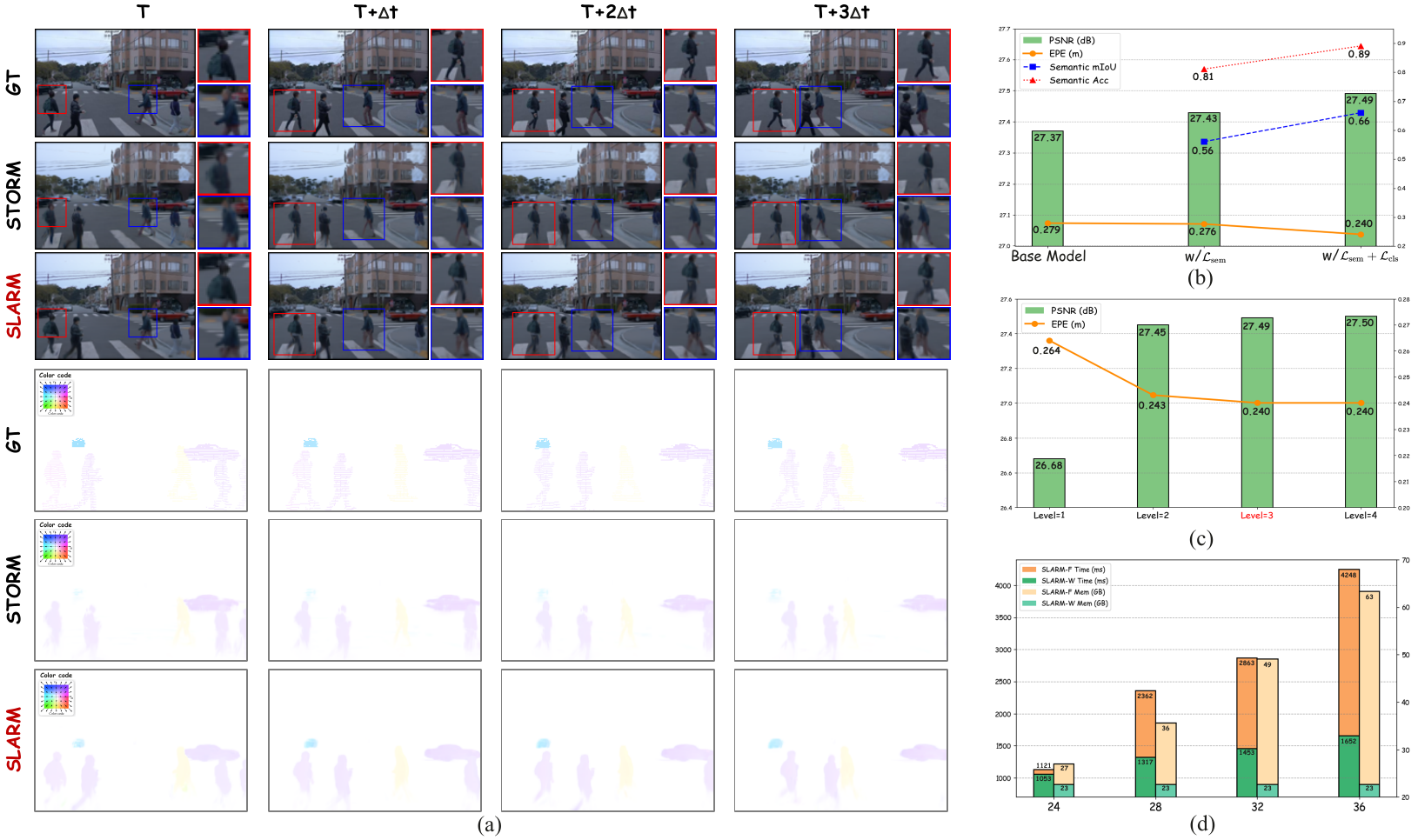}

    \vspace{-2mm}
    \caption{(a) Qualitative comparison of dynamic scenes.
        (b) Influence of different semantic loss terms on model performance.
        (c) Impact of varying motion levels on model performance.
        (d) Comparison of inference speed and memory usage between online and offline modes.}
    \label{fig:qualitative_1}
    \vspace{-5mm}
\end{figure*}

\subsection{Implementation Details}
\label{sec:implementation}

\noindent\textbf{Training.}
We train the model for 4 days on 64 Huawei Ascend 910B NPUs with a batch size of 64, using the AdamW~\cite{AdamW} optimizer to minimize the training loss over 200k iterations.

\vspace{0.5em}
\noindent\textbf{Supervision and Loss Functions.}
After aggregating the input frames (observations) across time into an amodal 4D scene representation, we can render the scene at any target timestamp within the observed temporal window. During training, we randomly sample multiple interpolation timestamps within this window for supervision. For each rendered frame, we optimize the following composite loss:
\vspace{-10pt}

\begin{equation}
    \begin{split}
        \mathcal{L}_{\mathrm{total}} ={} &
        \mathcal{L}_{\mathrm{rgb}} +
        \mathcal{L}_{\mathrm{depth}} +
        \lambda_{\mathrm{sky}} \, \mathcal{L}_{\mathrm{sky}}                                        \\
                                         & + \lambda_{\mathrm{reg}} \, \mathcal{L}_{\mathrm{reg}} +
        \lambda_{\mathrm{feat}} \, \mathcal{L}_{\mathrm{feat}},
    \end{split}
\end{equation}
where the individual loss terms are defined as follows.

The depth consistency loss enforces geometric alignment with the ground-truth depth map $\mathbf{D}_{\mathrm{gt}} \in \mathbb{R}^{H \times W}$:
\begin{equation}
    \mathcal{L}_{\mathrm{depth}} = \frac{1}{|\mathbf{M}_{\mathrm{valid}}|} \sum_{p \in \mathbf{M}_{\mathrm{valid}}} \bigl\| \hat{\mathbf{D}}_p - \mathbf{D}_{\mathrm{gt}, p} \bigr\|_1,
\end{equation}

\noindent where $\mathbf{M}_{\mathrm{valid}}$ is the set of valid pixel indices that have corresponding depth values in $\mathbf{D}_{\mathrm{gt}}$. Specifically, $\mathbf{D}_{\mathrm{gt}, p}$ and $\hat{\mathbf{D}}_p$ denote the ground-truth and predicted depth values at pixel $p$, respectively.

The sky regularization loss encourages transparency in sky regions by penalizing the predicted opacity values in the rendered alpha map $\hat{\mathbf{A}} \in \mathbb{R}^{H \times W}$. The sky mask $\mathbf{M}_{\mathrm{sky}}$ is obtained from DepthAnythingV2~\cite{depthanythingv2} by selecting pixels with zero depth. The loss is computed only over these sky pixels:
\begin{equation}
    \mathcal{L}_{\mathrm{sky}} = \frac{1}{|\mathbf{M}_{\mathrm{sky}}|} \sum_{p \in \mathbf{M}_{\mathrm{sky}}} \hat{\mathbf{A}}_p,
\end{equation}
where $\hat{\mathbf{A}}_p$ denotes the predicted alpha (opacity) at pixel $p$, and $|\mathbf{M}_{\mathrm{sky}}|$ is the number of sky pixels.

A motion regularization term suppresses higher-order coefficients under the prior that most scenes are static:
\begin{equation}
    \mathcal{L}_{\mathrm{reg}} = \sum_{l=0}^{3} \ \bigl\| \bm{m}_l \bigr\|_2^2.
\end{equation}

The feature alignment loss is given by:
\begin{equation}
    \mathcal{L}_{\mathrm{feat}} = \mathcal{L}_{\mathrm{sem}} \lor \mathcal{L}_{\mathrm{cls}},
\end{equation}
where $\lor$ denotes “or”. In our final training protocol, we first optimize the model using $\mathcal{L}_{\mathrm{sem}}$ as $\mathcal{L}_{\mathrm{feat}}$ for 200k iterations, followed by an additional 3k iterations with $\mathcal{L}_{\mathrm{cls}}$ substituted for $\mathcal{L}_{\mathrm{feat}}$. Specifically, the loss weights are set to $\lambda_{\mathrm{sky}} = 0.1$, $\lambda_{\mathrm{reg}} = 0.005$, and $\lambda_{\mathrm{feat}} = 1.0$ throughout training.

\begin{table*}[tb]
    \centering
    \begin{minipage}[t]{0.60\linewidth}
        \centering
        \caption{\textbf{Comparison to state-of-the-art methods on the WOD.} We compare photorealism and geometry metrics against generalizable feed-forward methods. PSNR, SSIM, and Depth RMSE (D-RMSE) are reported. SLARM-F denotes the model using full attention in offline mode, whereas SLARM-W uses window-attention in online mode. $^*$: reproduced by us. $^\dagger$: Non-sky region.}
        \label{tab:main_table}
        \resizebox{\linewidth}{!}{%
            \begin{tabular}{lccccccc}
                \toprule
                \multirow{2}{*}{Methods}            &
                \multicolumn{3}{c}{Dynamic-only}    & \multicolumn{3}{c}{Full image$^\dagger$}                                                                                                                                 \\
                                                    & PSNR$\uparrow$                           & SSIM$\uparrow$          & D-RMSE$\downarrow$
                                                    & PSNR$\uparrow$                           & SSIM$\uparrow$          & D-RMSE$\downarrow$                                                                                  \\
                \midrule
                LGM~\citep{lgm}                     & 17.36                                    & 0.216                   & 11.09                  & 18.53                   & 0.447                   & 9.07                   \\
                LGM*~\citep{lgm}                    & 19.58                                    & 0.443                   & 9.43                   & 23.59                   & 0.691                   & 8.02                   \\
                GS-LRM$^*$~\citep{gs-lrm}           & 20.02                                    & 0.520                   & 9.95                   & 25.18                   & 0.753                   & 7.94                   \\
                MapAnything~\citep{mapanything} & -                                        & -                       & 20.99                  & -                       & -                       & 13.53                  \\
                STORM$^*$~\citep{storm}             & \cellcolor{third}22.03                   & \cellcolor{third}0.623  & \cellcolor{third}7.50  & \cellcolor{third}25.86  & \cellcolor{third}0.804  & \cellcolor{third}5.47  \\
                \midrule
                \multicolumn{4}{l}{\textit{Ours}}                                                                                                                                                                              \\
                SLARM-W                             & \cellcolor{second}23.20                  & \cellcolor{second}0.676 & \cellcolor{second}6.38 & \cellcolor{second}27.30 & \cellcolor{second}0.825 & \cellcolor{second}4.75 \\
                SLARM-F                             & \cellcolor{best}{23.51}                  & \cellcolor{best}{0.691} & \cellcolor{best}{6.16} & \cellcolor{best}{27.49} & \cellcolor{best}{0.828} & \cellcolor{best}{4.57} \\
                \bottomrule
            \end{tabular}%
        }
    \end{minipage}%
    \hfill
    \begin{minipage}[t]{0.36\linewidth}
        \centering
        \caption{\textbf{Quantitative comparison of semantic segmentation performance}. Our method {SLARM} achieves the best mIoU and accuracy among all methods.}
        \label{tab:semantic_segmentation}
        \resizebox{\linewidth}{!}{%
            \begin{tabular}{lcc}
                \toprule
                Method                                      & mIoU $\uparrow$            & Acc $\uparrow$             \\ \hline
                EfficientViT-Seg~\cite{cai2023efficientvit} & 0.4352                     & 0.7637                     \\
                Mask2Former-R50~\cite{cheng2022masked}      & 0.4429                     & 0.7082                     \\
                SegMAN~\cite{fu2025segman}                  & 0.4567                     & 0.7186                     \\
                SegFormer~\cite{xie2021segformer}           & 0.4660                     & 0.7572                     \\
                OffSeg-B~\cite{zhang2025revisiting}         & 0.4612                     & 0.7417                     \\
                OffSeg-L~\cite{zhang2025revisiting}         & 0.4868                     & 0.7635                     \\
                LSeg~\cite{li2022lseg}                      & \cellcolor{third}{0.4876}  & \cellcolor{third}{0.7976}  \\
                Mask2Former-Swin~\cite{cheng2022masked}     & \cellcolor{second}{0.5505} & \cellcolor{second}{0.8192} \\ \hline
                {SLARM}                                     & \cellcolor{best}{0.6663}   & \cellcolor{best}{0.8923}   \\
                \bottomrule
            \end{tabular}%
        }
    \end{minipage}
    \vspace{-3mm}
\end{table*}

\section{Experiments}
\label{sec:experiments}
\noindent\textbf{Dataset.}
We conduct our main experiments on the Waymo Open Dataset (WOD)~\cite{waymo_open_dataset}, a large-scale autonomous driving benchmark featuring rich dynamics and synchronized multi-camera video sequences. Our experiments use 1,000 driving sequences---798 for training and 202 for validation---each approximately 20 seconds long at 10~fps (yielding $\sim$200 frames per sequence). Input images are downsampled by a factor of 8 to $160 \times 240$. To align the resolution of the extracted LSeg~\cite{li2022lseg} features with it, image inputs for feature extraction use $320 \times 480$.

We present main experimental results and analyses below. More experimental results analyses are provided in the \textcolor{red}{\textbf{Supplementary Material}}.

\subsection{Dynamic Reconstruction}
\label{sec:4.1}

\noindent\textbf{Setup and Baselines.}
We report standard metrics: PSNR and SSIM for photometric fidelity, and RMSE for depth accuracy. We compare SLARM with existing generalizable feed-forward reconstruction models on the validation split of the WOD~\cite{waymo_open_dataset}. To better understand model behavior in dynamic environments, we further analyze performance separately on the \textit{full image} and on \textit{dynamic regions only} (masked using ground-truth motion annotations).

\noindent\textbf{Results.}
Quantitative results are summarized in Table~\ref{tab:main_table}. SLARM outperforms all generalizable feed-forward methods across all metrics, with consistent gains of 1.6 dB in PSNR on full images and improvements of over 1.5 dB in PSNR and 0.07 in SSIM on dynamic regions—highlighting its superior photorealism and geometric accuracy.

\subsection{Flow Estimation}
\label{sec:4.2}

\noindent\textbf{Setup and Baselines.}
We evaluate flow estimation on the validation split of the WOD~\cite{waymo_open_dataset}, which provides ground-truth 3D scene flows. Following the protocol of STORM, we report standard metrics including End-Point Error in 3D (EPE3D), $\text{Acc}_{\text{5}}$, $\text{Acc}_{\text{10}}$, and angular error $\theta_{\text{err}}$. Notably, while STORM originally evaluates only on input context frames, we adopt a more comprehensive evaluation strategy by considering the entire video sequence—including both input context frames and output interpolated target frames—to enable a fairer and more objective comparison.

\noindent\textbf{Results.}
As shown in Table~\ref{table:epe}, SLARM outperforms all competing methods across all metrics, achieving significant improvements particularly in EPE3D and angular error $\theta_{\text{err}}$. This demonstrates the importance of higher-order motion modeling for dynamic scene reconstruction. In contrast to STORM, which assumes uniform motion for all scene objects, our method can capture more complex motion patterns, leading to more accurate 3D scene flow predictions.

\begin{figure*}[t]
    \centering
    \includegraphics[width=1.0\linewidth]{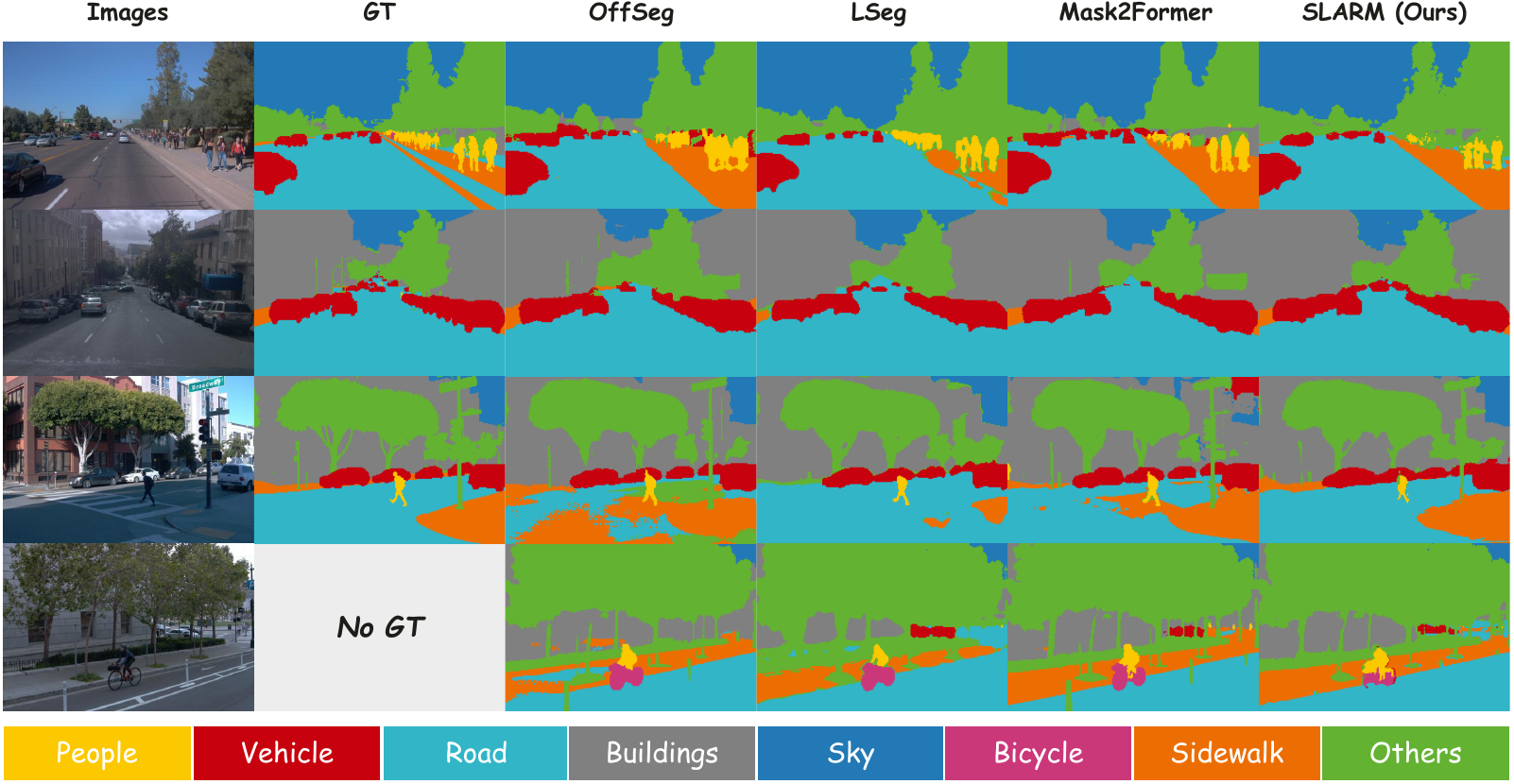}

    \vspace{-2mm}
    \caption{Qualitative comparison of semantic segmentation performance. Our method produces more accurate and coherent segmentations compared to previous 2D segmentation approaches.}
    \label{fig:semantic_1}
    \vspace{-5mm}
\end{figure*}

\subsection{Semantic Segmentation in 3D Reconstruction}
\label{subsec:semantic_segmentation}
\noindent\textbf{Setup and Baselines.}
Following the same protocol as Sec.~\ref{sec:4.1}, we use semantically annotated validation images from WOD~\cite{waymo_open_dataset} for semantic segmentation evaluation, with standard mIoU and Acc metrics. We compare our method with several SOTA methods, all of which are pre-trained on Cityscapes~\cite{cordts2016cityscapes} and spanning diverse cutting-edge paradigms. To ensure a fair comparison, all methods are evaluated at the same image resolution of $160 \times 240$.

\noindent\textbf{Results.}
Quantitative results are summarized in Table~\ref{tab:semantic_segmentation}. SLARM achieves highly coherent and accurate semantic predictions within the 3D reconstruction pipeline, and outperforms the strong 2D baselines.
We attribute this improvement to two key factors: (i) the combination of semantic priors inherited from LSeg and semantic knowledge learned from ground truth, and (ii) the explicit incorporation of 3D geometric priors, which substantially enhance the expressiveness and robustness of learned features.

\subsection{Qualitative Results}
\label{sec:4.3}

\noindent\textbf{Dynamic Reconstruction Results.}
We evaluate on a challenging walking-person scene, as shown in Figure~\hyperref[fig:qualitative_1]{\ref*{fig:qualitative_1}(a)}, where the horizontal axis denotes timesteps. The top three rows show ground-truth RGB, STORM's reconstruction, and ours. Red/blue boxes highlight two pedestrians, with close-ups for detail.
STORM assumes uniform velocity, yielding nearly identical poses across frames and failing to capture natural gait dynamics. In contrast, our higher-order motion modeling accurately reconstructs time-varying pose changes. This is further confirmed by 3D scene flow (bottom three rows): our method captures fine-grained motion variations in speed and direction that STORM misses.

\noindent\textbf{Semantic Segmentation Results.}
As shown in Table~\ref{tab:semantic_segmentation} and Figure~\ref{fig:semantic_1}, SLARM achieves state-of-the-art semantic segmentation with highly coherent results. We evaluate on common autonomous driving scenarios and key categories, showing strong accuracy on critical classes (e.g., people, vehicle, road, sidewalk) as well as the ``others'' category—covering safety-relevant objects.
All results are obtained by matching language-aligned features with CLIP~\cite{clip} text embeddings. These semantically rich features can further enhance a large language model's understanding of dynamic scenes.

\noindent\textbf{Streaming Inference Results.}
As shown in Figure~\hyperref[fig:qualitative_1]{\ref*{fig:qualitative_1}(d)}, SLARM-W outperforms SLARM-F in inference speed and memory consumption. With window attention, it achieves linear inference time and stable memory usage, making it well-suited for long-sequence streaming inference.

\subsection{Ablation Study}
\label{sec:4.4}

\noindent\textbf{High-order motion representation.}
As introduced in Sec.~\ref{sec:3.2}, we model dynamics using a high-order motion representation, where increasing orders correspond to higher kinematic derivatives (e.g., velocity, acceleration, jerk).  As shown in Figure~\hyperref[fig:qualitative_1]{\ref*{fig:qualitative_1}(c)}, we find that a \textbf{3rd-order} model suffices for optimal performance: over short time intervals, real-world motion is well-approximated by jerk-level dynamics, with higher orders yielding diminishing returns.

\noindent\textbf{Enhancing dynamic reconstruction with semantic guidance.}
We leverage semantic consistency as a temporal regularizer: objects with coherent semantic identity (e.g., cars, pedestrians) should follow smooth, physically plausible trajectories. As shown in Figure~\hyperref[fig:qualitative_1]{\ref*{fig:qualitative_1}(b)} experiments confirm its benefit: $\mathcal{L}_{\text{sem}}$ yields a lower Flow EPE metric, and the subsequent adoption of $\mathcal{L}_{\text{cls}}$ further reduces it. Furthermore, with improved dynamic performance, PSNR and semantic metrics are correspondingly improved.

\begin{table}[t]
    \centering
    \caption{\textbf{Comparison of scene flow estimation on the WOD.}}
    \label{tab:scene_flow_estimation}
    \scriptsize
    \begin{tabular}{lccccc}
        \toprule
        Methods                         & EPE $ (m) \downarrow$    & $\text{Acc}_{5} (\%)$ $\uparrow$ & $\text{Acc}_{10} (\%) \uparrow$ & $\theta$ (rad) $\downarrow$ \\
        \midrule
        NSFP ~\cite{li2021nsfp}         & 0.703                    & 41.03                            & 52.16                           & 0.934                       \\
        NSFP++ ~\cite{najibi2022motion} & 0.724                    & 51.56                            & 60.04                           & 1.103                       \\
        STORM~\cite{storm}              & \cellcolor{second}0.304  & \cellcolor{second}79.01          & \cellcolor{second}83.74         & \cellcolor{second}0.667     \\
        \midrule
        \multicolumn{4}{l}{\textit{Ours}}                                                                                                                             \\

        SLARM-F                         & \cellcolor{best}{0.240}  & \cellcolor{third}{78.15}         & \cellcolor{third}{83.08}        & \cellcolor{best}{0.540}     \\
        SLARM-W                         & \cellcolor{third}{0.337} & \cellcolor{best}{81.07}          & \cellcolor{best}{84.26}         & \cellcolor{third}{0.725}    \\
        \bottomrule
    \end{tabular}
    \vspace{-5mm}
    \label{table:epe}
\end{table}

\section{Conclusion}
In this work, we present \textbf{S}treaming and \textbf{L}anguage-\textbf{A}ligned \textbf{R}econstruction \textbf{M}odel (SLARM), a scalable feed-forward 4D Gaussian Splatting framework that jointly recovers 3D scene flow, metric depth, and language-aligned semantics from posed video. Trained purely via rendering-based self-supervision, SLARM captures complex motion without ground-truth flow and supports real-time incremental inference. The model supports language-aligned semantics, enabling integration with Vision-Language Models (VLMs) for Vision–Language–Action (VLA) systems. However, SLARM currently requires accurate camera poses and struggles with complex materials like glass or mirrors due to its reliance on photometric consistency. Future work will explore self-calibration and more realistic scene representations to address these limitations.
{
  \small
  \bibliographystyle{ieeenat_fullname}
  \bibliography{main}
}

\clearpage
\setcounter{page}{1}
\maketitlesupplementary

\section{More Implementation Details}
\label{sec: Implementation}

\noindent\textbf{Model architecture.}
As the standard configuration, our model employs a 12-layer Alternating-Attention Transformer~\cite{vggt}, which interleaves frame-wise and global self-attention mechanisms. Each attention layer operates with a feature dimensionality of 768. The input image is processed using a Vision Transformer (ViT) with a patch size of $8 \times 8$, yielding a sequence of image tokens that serve as input to a Gaussian Decoder. This decoder comprises three lightweight MLP-based task-specific heads: a \textit{Gaussian head}, a \textit{motion head}, and a \textit{semantic head}.

The Gaussian head regresses the geometric and appearance parameters of 3D Gaussians, specifically the pixel-aligned depth $d \in \mathbb{R}$, rotation represented by a unit quaternion $\bm{q} \in \mathbb{R}^4$, scale $\bm{s} \in \mathbb{R}^3$, opacity $\alpha \in [0,1]$, and color $\bm{c} \in \mathbb{R}^3$, collectively forming a 12-dimensional parameter vector per Gaussian primitive.

Each Gaussian primitive is parameterized by position, scale, rotation (quaternion), opacity, RGB color, and an auxiliary depth value. We use the following activation functions to map raw network outputs to valid physical ranges:

\begin{itemize}
    \item \textbf{Scale}: $\text{scale} = \min\left(\exp(x + \texttt{scale\_offset}),\ \texttt{0.5}\right)$,
          where $\texttt{scale\_offset} = -0.693$ (i.e., $\log(0.5)$). This initialization biases the model to start from relatively large Gaussians and shrink during training, which we empirically find beneficial for stable self-supervised learning of motion.

    \item \textbf{Opacity}: $\sigma = \texttt{sigmoid}(x - 2.0)$, following GS-LRM~\cite{gs-lrm}, which encourages sparse initialization (low opacity) and reduces floaters.

    \item \textbf{RGB}: $\mathbf{c} = \texttt{sigmoid}(x)$, clamping colors to $[0, 1]$.

    \item \textbf{Depth}: $d = \texttt{near} + \texttt{sigmoid}(x) \cdot (\texttt{far} - \texttt{near})$,
          with $\texttt{near} = 0.2$ and $\texttt{far} = 400$, ensuring depth values lie within a physically plausible range.

    \item \textbf{Quaternion}: No activation is applied.
\end{itemize}

The motion head predicts 12-dimensional third-order motion properties. For each order $l \in \{1,2,3\}$, it outputs a scalar velocity magnitude and a 3-dimensional directional vector, resulting in $4 \times 3 = 12$ dimensions.

The semantic head produces a 64-dimensional semantic feature map intended for novel-view feature rendering. This feature map is subsequently refined by an auxiliary MLP decoder that expands its dimensionality from 64 to 512, ensuring compatibility with the LSeg feature space.

\section{Why We Choose LSeg}
\label{sec:semantic}

Within the framework of feature distillation, the capability of teacher features is crucial for the final semantic reconstruction performance of the model. In order to enable our features to possess language alignment capability, we select three types of CLIP-related features and conduct experimental comparisons.

\begin{figure*}[!t]
    \centering
    \includegraphics[width=1.0\linewidth]{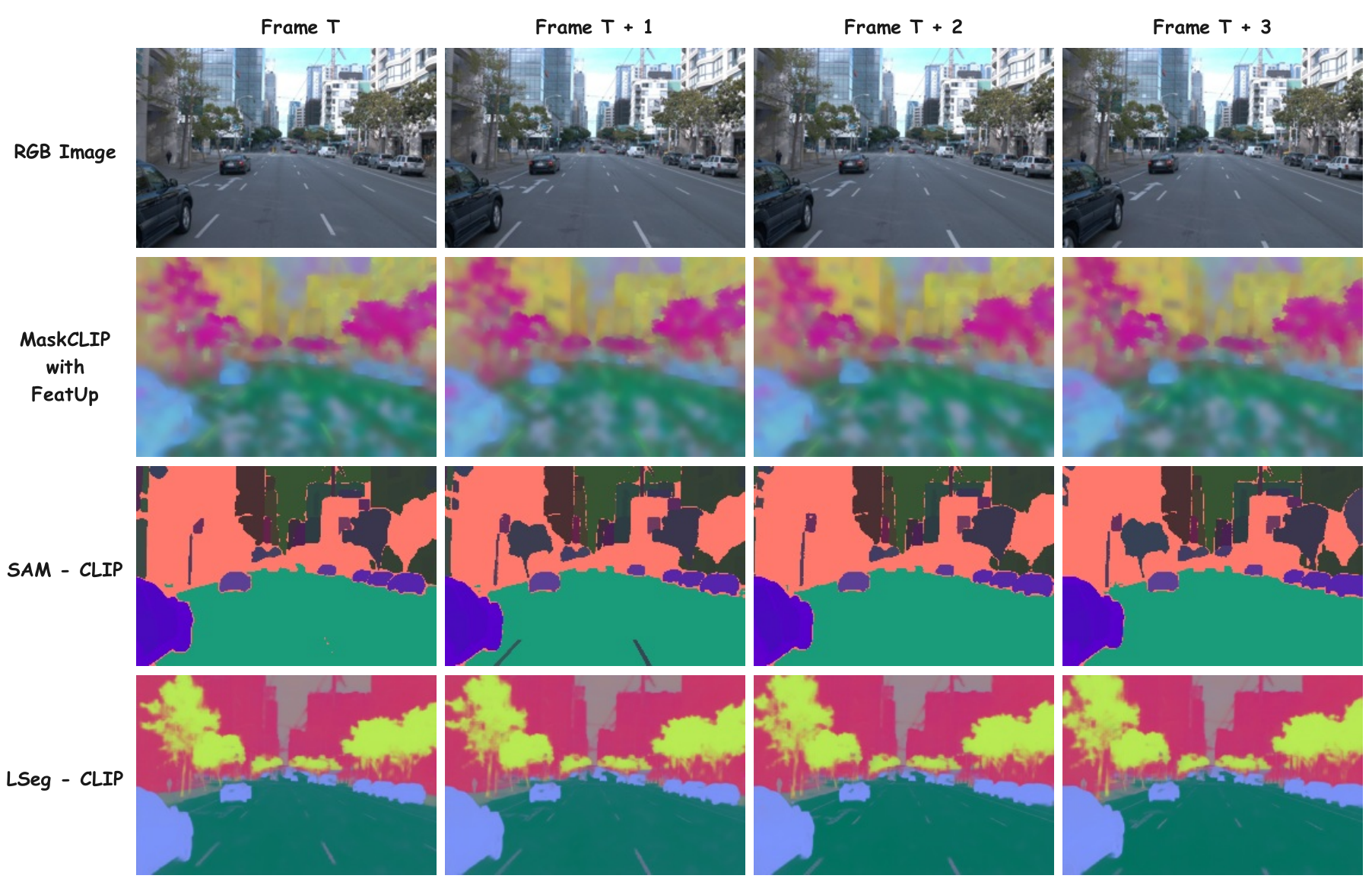}

    \caption{Comparison of different language-aligned features. The first row is the input consists of a set of images from adjacent frames, the next three rows correspond to the three types of features for each frame, respectively.}
    \label{fig:feature_comparison1}
\end{figure*}

\noindent \ding{113} \textbf{MaskCLIP}~\cite{dong2023maskclip}: Standard CLIP~\cite{clip} computes similarity only between text and global visual feature during contrastive learning, resulting in the local visual features not being strictly aligned with text. MaskCLIP is a CLIP variant that achieves alignment between local features and text. However, its features undergo significant spatial downsampling, yielding semantically condensed but geometrically distorted representations. In SAB3R~\cite{chen2025sab3r}, MaskCLIP features are processed via FeatUp~\cite{fu2024featup}, which upsamples the low-resolution features to restore geometric fidelity. As shown in the second row of Figure~\ref{fig:feature_comparison1}, in our experiments, this upsampling method can accurately restore the edges of some instances. However, it tends to cause feature confusion, which easily interferes with feature learning.

\noindent \ding{113} \textbf{SAM-CLIP}: We use SAM-CLIP to denote instance-level CLIP features obtained by extracting CLIP features from SAM-segmented regions. Similar to PE3R~\cite{pe3r}, we use SAM to segment images and extract CLIP features from the segmented regions. Meanwhile, we employ SAM2~\cite{ravi2024sam} to perform instance ID alignment across different frames and views, and conduct feature aggregation for identical instances. As shown in the third row of Figure~\ref{fig:feature_comparison1}, benefiting from the segmentation prior, the instance boundaries in the SAM-CLIP feature map are extremely clear. However, the segmentation prior also introduces several drawbacks, such as the presence of empty feature regions (where no instances are segmented) and potential feature jumps across frames (resulting from jumps in instance segmentation results across frames). These drawbacks make it less suitable for 4D reconstruction that involves a temporal dimension. Additionally, the complex processing procedure leads to very low efficiency in extracting such features.

\noindent \ding{113} \textbf{LSeg-CLIP}~\cite{li2022lseg}: LSeg-CLIP extends CLIP to the semantic segmentation task and enables text query-driven semantic segmentation. Specifically, LSeg-CLIP calculates the similarity between visual features and text features of various categories, which serves as the basis for category classification. Additionally, LSeg-CLIP adopts a visual encoder with low-magnification downsampling, allowing its features to retain a certain degree of geometric structure. As shown in the  last row of Figure~\ref{fig:feature_comparison1}, in our experiments, LSeg-CLIP features exhibit strong semantic expression capabilities, with sufficiently unified semantics for each category. Although its instance boundaries are less clear than those of SAM-CLIP, it possesses advantages that SAM-CLIP lacks: the absence of feature-less regions, the continuity between inter-frame features, and the efficiency of feature extraction. After comprehensive comparison, we adopt LSeg-CLIP features as our teacher features to endow Gaussian primitives with the capability of semantic reconstruction.

\section{Results in more outdoor datasets}
We extend SLARM to two more large-scale driving datasets—NuScenes~\cite{caesar2020nuscenes} and Argoverse2~\cite{wilson2argoverse}. As shown in Table~\ref{tab:dataset_performance}, SLARM consistently outperforms STORM in dynamic scene reconstruction.

\begin{table}[tb]
    \centering
    \caption{Comparison on NuScenes and Argoverse2 Datasets.}
    \vspace{-3mm}
    \label{tab:dataset_performance}
    \small 
    \setlength{\aboverulesep}{0.5ex}  
    \setlength{\belowrulesep}{0.5ex}
    \begin{tabular*}{\columnwidth}{@{\extracolsep{\fill}}lcccc@{}}
        \toprule
        \multirow{2}{*}{Method} & \multicolumn{2}{c}{NuScenes} & \multicolumn{2}{c}{Argoverse2} \\
        \cmidrule(lr){2-3} \cmidrule(lr){4-5}
        & PSNR  $ \uparrow $  & D-RMSE  $ \downarrow $  & PSNR  $ \uparrow $  & D-RMSE  $ \downarrow $  \\
        \midrule
        STORM     & 26.25 & 4.78 & 26.13 & 9.04 \\
        SLARM-F   & \textbf{26.71} & \textbf{4.24} & \textbf{26.49} & \textbf{8.65} \\
        SLARM-W   & \underline{26.42} & \underline{4.29} & \underline{26.26} & \underline{8.71} \\
        \bottomrule
    \end{tabular*}
    \vspace{-4mm}
\end{table}

\section{More Experiment Results}

\begin{figure*}[b]
    \centering
    \includegraphics[width=0.88\linewidth]{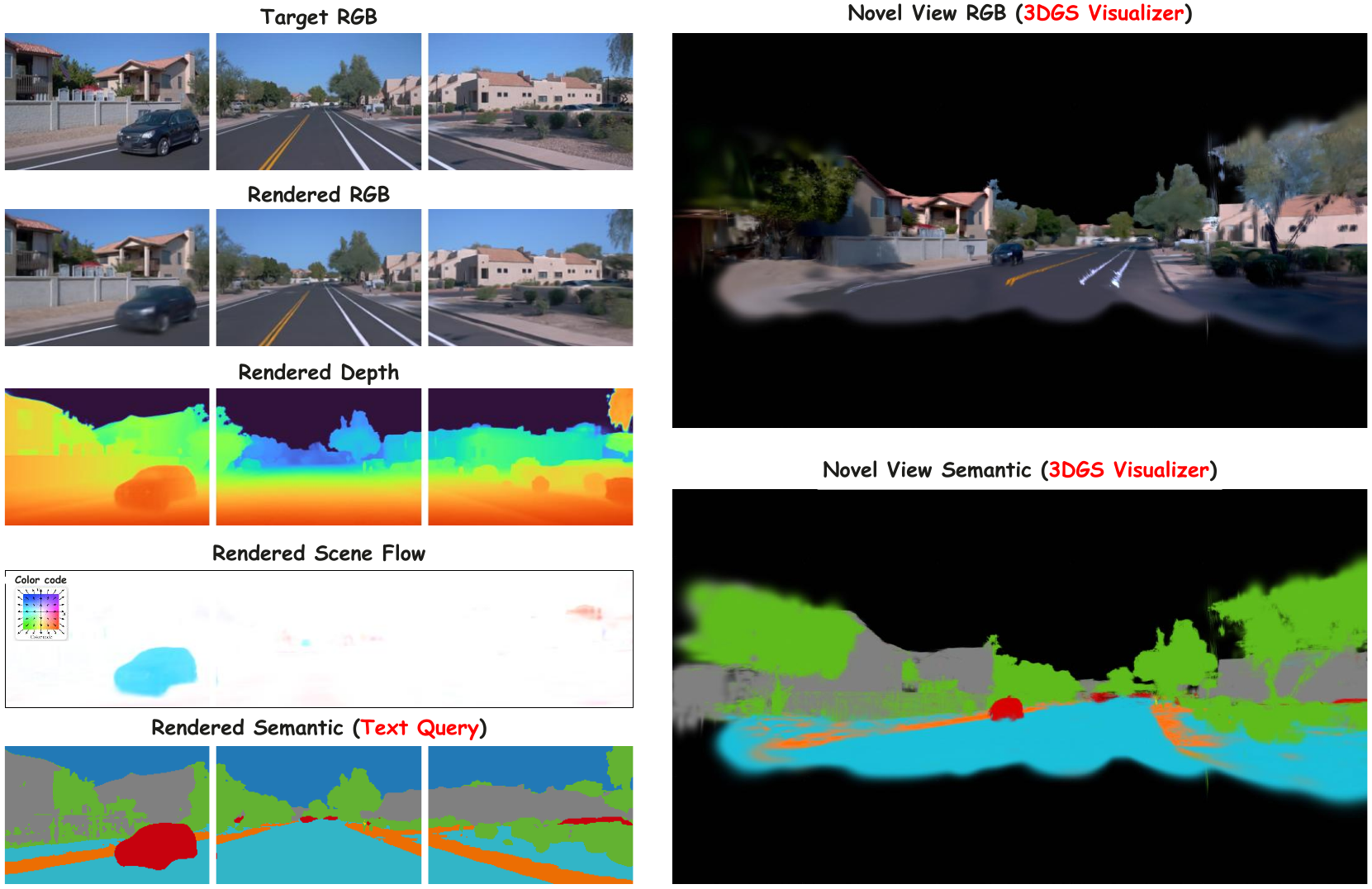}
  
    \caption{
    Qualitative results on a simple outdoor scene: left shows rendered RGB, depth, 3D scene flow, and semantic map from predicted 4DGS; right displays a novel view of the 4DGS in a 3DGS visualizer.}
    \label{fig:23}
  \end{figure*}

\begin{figure*}[t]
    \centering
    \includegraphics[width=0.88\linewidth]{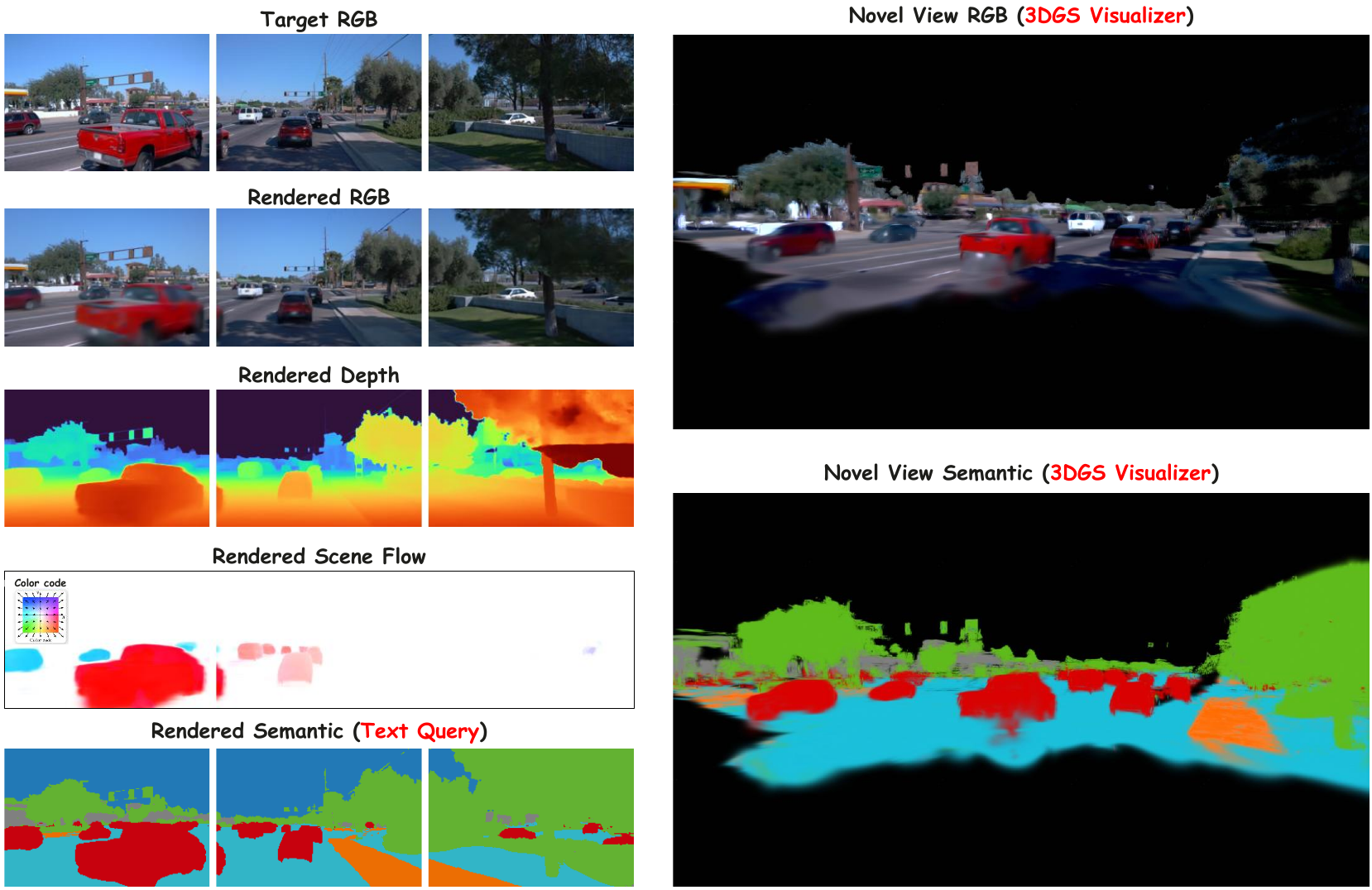}
  
    \caption{
    Qualitative results on a complex outdoor scene: left shows rendered RGB, depth, 3D scene flow, and semantic map from predicted 4DGS; right displays a novel view of the 4DGS in a 3DGS visualizer.}
    \label{fig:25}
  \end{figure*}

\begin{figure*}[t]
    \centering
    \includegraphics[width=0.88\linewidth]{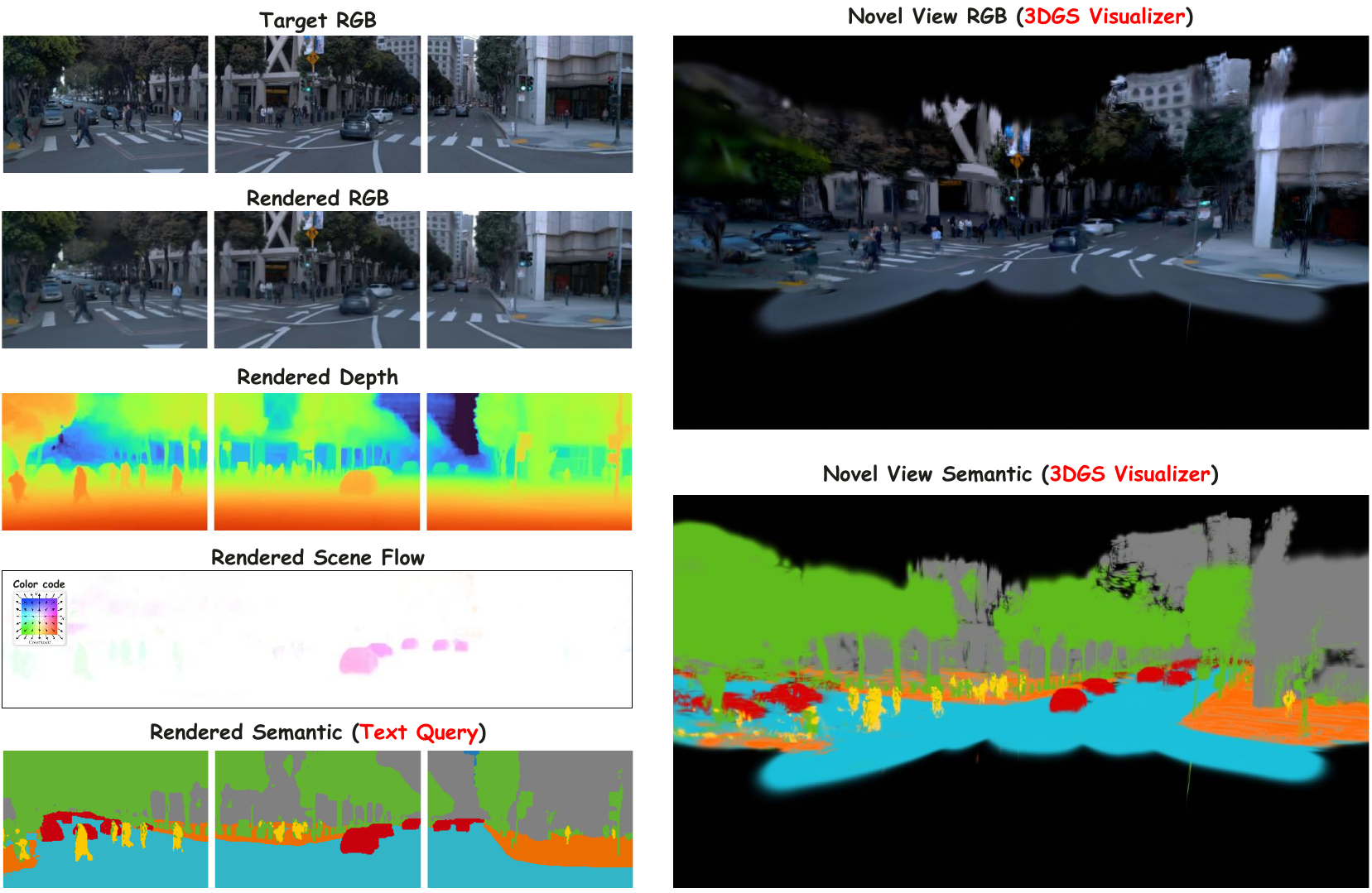}
  
    \caption{
        Qualitative results on a complex outdoor scene: left shows rendered RGB, depth, 3D scene flow, and semantic map from predicted 4DGS; right displays a novel view of the 4DGS in a 3DGS visualizer.}
    \label{fig:27}
  \end{figure*}

\begin{figure*}[t]
    \centering
    \includegraphics[width=0.88\linewidth]{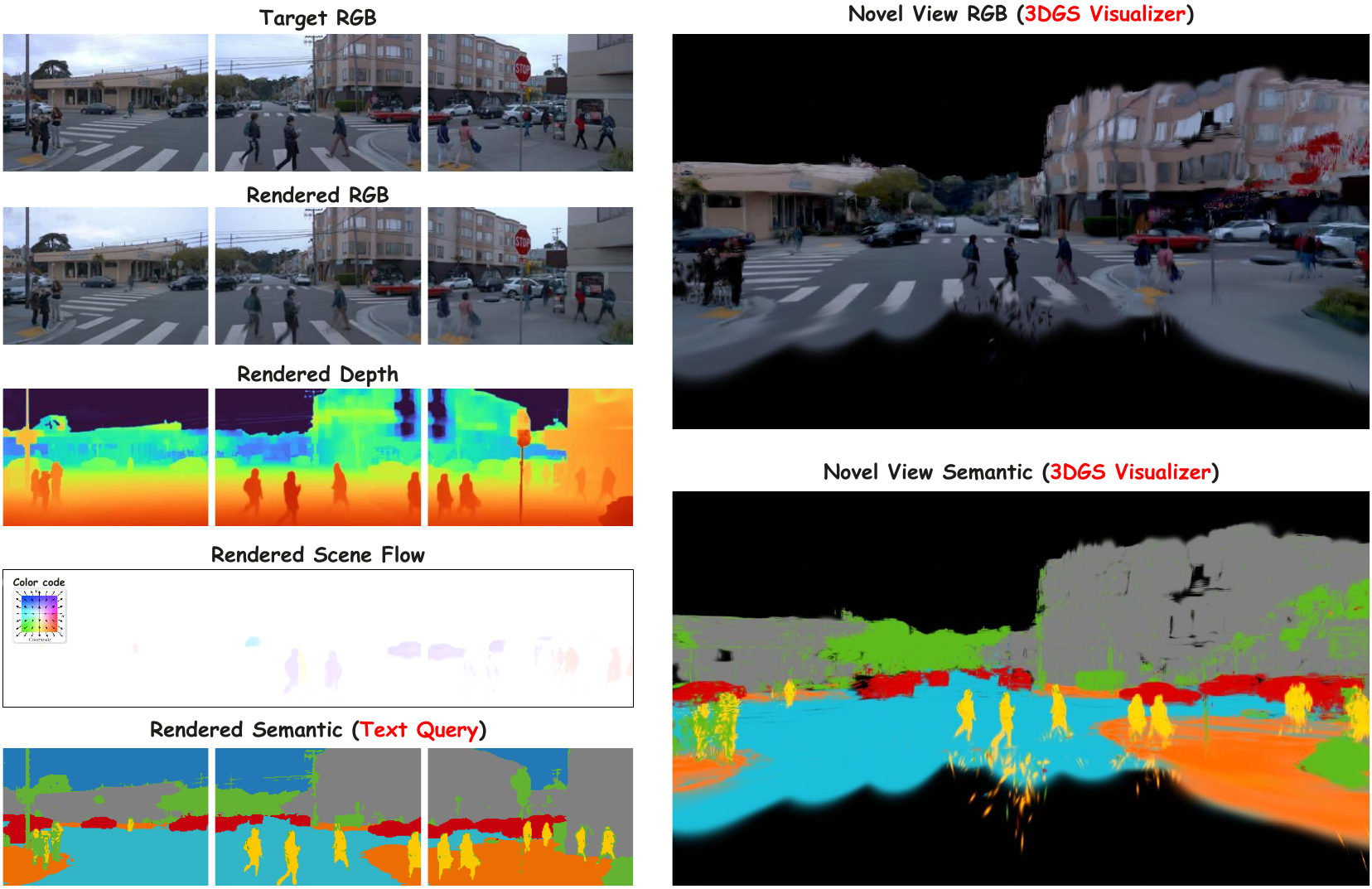}
  
    \caption{
    Qualitative results on a complex outdoor scene: left shows rendered RGB, depth, 3D scene flow, and semantic map from predicted 4DGS; right displays a novel view of the 4DGS in a 3DGS visualizer.}
    \label{fig:29}
  \end{figure*}

We present additional image and video results captured from novel viewpoints across a diverse set of dynamic scenarios. These include relatively simple cases—such as scenes with sparse moving objects exhibiting smooth motion (see Figure~\ref{fig:23})—as well as highly complex environments characterized by multiple simultaneously moving people and heterogeneous dynamic objects (see Figures~\ref{fig:25}--\ref{fig:29}). Further supplementary results, provided in the accompanying folder, corroborate that our method, SLARM, consistently delivers robust and high-fidelity reconstructions across this spectrum of scene complexity. Notably, SLARM preserves strong temporal coherence, accurate geometry, and photorealistic detail under both subtle motions and intricate multi-agent interactions, underscoring its generalizability and practical efficacy in real-world dynamic settings.

Additional videos results are available in our page \href{https://kevinchiu19.github.io/SLARM/}{https://kevinchiu19.github.io/SLARM/}.

\end{document}